# Accurate shape and phase averaging of time series through Dynamic Time Warping

**George Sioros*[1] & Kristian Nymoen[1]**

[1]RITMO Centre for Interdisciplinary Studies in Rhythm, Time and Motion
University of Oslo

*correspondence: georgios.sioros@imv.uio.no; georgios.sioros@fe.up.pt

**Abstract**

Time series averaging is a crucial part of series summarization with important applications in data mining, signal noise suppression and information retrieval. Current averaging methods employ dynamic time warping (DTW) to correct for nonlinear phase variation between the sequences. They consider phase differences to be accidental and reduce durational information to sequential. Thus, they introduce time distortion that hinders the accurate estimation of the mean shape of the signals, its distinctive durational features, and the mean location of landmarks, all of which are required in many summarization applications. To address this problem, we propose a novel averaging method which preserves durational information due to a simple conversion of the output of DTW into a time sequence and an innovative iterative averaging process. We show that our method accurately estimates the ground truth mean sequences and mean temporal location of landmarks in synthetic and real-world datasets and outperforms state-of-the-art methods.

## 1 Introduction

As time is a fundamental dimension of any physical or biological system, data collected in many domains of application often appear as time-dependent sequences of observations, that is, as time series. Analysis of collection of time series may be challenging when the phase between them varies nonlinearly. Dynamic Time Warping (DTW) is a well-known method that associates similar features between two time series even their temporal dynamics differ. From its early days in speech recognition [1,2], DTW has found a wide range of applications, from general time series summarization [3] and clustering [4,5] to specific applications such as music and motion information retrieval [6] or signature verification [7,8], and has seen various improvements [9–11].

DTW is commonly used to correct accidental differences in the relative phase between signals prior to measuring their amplitude difference. However, in many applications in which timing is a critical factor, phase variation between the signals is more than a mere artifact of the data collection process. It is rather a main point of interest that needs to be measured and assessed [12–14]. Moreover, in various fields of research, multiple time series need to be summarized through descriptive statistics. Current methods of time series summarization and averaging are aimed at data mining and classification tasks [3,15–19]. They are based on aligning the time series through a DTW algorithm [20], a process which reduces temporal information to sequential, that is, while the order of time points is preserved, exact durational information is lost as several time points may be collapsed into a single one or a single point repeated many times. Thus, the resulting mean sequences do not effectively average the phase variation and do not capture salient temporal characteristics of the signals, such as mean locations and durations of their features and landmarks. Consequently, more common methodologies for processing phase variation independently of amplitude variation have been the detection and statistical processing of events and onsets [14,21], and Functional Data Analysis (FDA) [12,22,23]. The former method requires defining specific landmarks of interest which may not always be easy to detect and, in any case, do not capture continuous phase variations. The later, FDA, requires a good understanding of the signals and high level of expertise to achieve good results and to determine and fine tune appropriate parameters for each dataset [12]. Therefore, either technique is difficult to automate and does not offer a generic methodology. Furthermore, although DTW has been employed to study lead-lag structures between time series, such applications have been limited, for example to single pairs of financial time series [24,25]. This is mainly because a generic method for the quantification of local phase variation between the signals of a collection, which makes it easy to extract this information from the output of DTW and process it statistically, is still missing.

This paper addresses the challenge of averaging phase variation in a set of sequences—or generally domain variation for signals defined in other domains besides time—to produce a mean sequence which accurately reproduces the mean temporal locations and durations of their features. Through a simple transformation of the optimal alignment between two sequences, such as the one produced by DTW, we compute what we call the Time Warp Profile (TWP), that is, a time sequence that measures the local phase differences between two sequences along a fixed time scale. Based on the TWP we propose a method for the computation of a mean sequence from a collection of sequences that averages the local phase differences between them. Additionally, to describe the statistical distribution of the signals in time the standard deviation around that mean may be computed.

To define a mean sequence, a similarity measure is needed to quantify the distance between a pair of sequences. The DTW distance has been previously used for this purpose [3,20]. However, DTW is focused on amplitude differences rather than phase differences, which is the focus of this paper. Recently the Time Alignment Measure (TAM) has been proposed [26] as a measure of the amount of warping between two sequences produced by DTW. It quantifies the amount of time distortion between the sequences such that two signals with a constant phase difference would be deemed as highly similar, even if their phase difference were large, as their shapes are not deformed. In this sense, TAM measures the similarity of the *shape* of two sequences in time, but not their relative phase. To address the lack of a consistent measure of phase difference and facilitate the definition of a corresponding mean sequence, we propose a new measure, the *Phase* distance, based on the TWP between two sequences.

The rest of the document is structured as follows. We present important details of DTW and related work in section 2, the main theoretical framework of TWP in section 3, the new averaging method in section 4, and, finally, in section 5, an evaluation of our averaging method with synthetic and real-word datasets and a comparison against the available ground truth and state-of-the-art methods.

## 2    Related work

In this section, we introduce the main concept of DTW, provide relevant definitions, and discuss the main limitations of current time series averaging methods related to the problem of averaging phase variation.

A direct comparison of two time series associates their elements based on their position in the sequences alone. For two sequences $S_1$ and $S_2$ of lengths $n$ and $m$ respectively, time point $i$ of sequence $S_1$ is associated with the time point at the same position $i$ in $S_2$. Such a comparison is sensitive to subtle, often accidental phase differences between otherwise similar sequences. DTW re-aligns the sequences to better match similar features by creating associations between elements that do not necessarily correspond to the same temporal position (Figure 1). The aim of the DTW algorithm is to associate the most similar elements between the two sequences while preserving their order. To this end, a similarity matrix $M(m \times n)$ is first constructed that contains the distance between all pairs of elements. The two dimensions of the similarity matrix are the time scales of the two sequences. Different distance metrics may be used in the construction of the matrix. The most common one is the Euclidean distance.

The DTW associations take the form of a warping path $W$ consisting of steps through the similarity matrix that move forward in both dimensions. Different step patterns may be used [1,27], but the most common is single forward steps, either horizontal, vertical or diagonal, which has the advantage of not omitting any elements of either sequences. Each step in the similarity matrix introduces a cost to the warping path which is equal to the distance between the corresponding elements. Having defined the similarity matrix $M$ and step pattern, the DTW algorithm finds the optimal warping path $W$ by minimizing the total cost of stepping through $M$. To reduce the computational complexity, dynamic programming is employed.

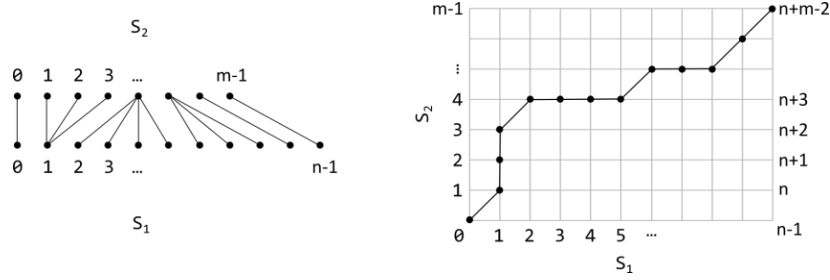

*Figure 1. Example of a warping path, visualized as associations between the elements of two sequences S₁ and S₂ (left), and as steps through their similarity matrix (right).*

Global constraints are commonly introduced to the warping path. The most common requirement is that the first and last elements of one sequence are associated with the respective elements of the other. Other constraints aim at reducing the search space to a certain part of the similarity matrix; for instance, the Itakura Parallelogram [28] and Sakoe-Chiba Band [1] are two examples of constraining the warping path around the diagonal.

A warping path may be formally expressed as a set of associations $W$ between the indices $t_1 \in \{0, \ldots, m-1\}$ and $t_2 \in \{0, \ldots, n-1\}$ of the two sequences $S_1$ and $S_2$:

$$\begin{aligned} W &= (w_1, w_2, \ldots w_k) \\ w_i &= (t_1, t_2)_i, w_0 = (0,0), w_k = (m-1, n-1) \\ &\max(m, n) \leq k \leq m+n-1 \\ \Delta w_i &= (w_{i+1} - w_i) \in \{(1,1), (1,0), (0,1)\} \end{aligned}$$

Equation 1

Since the warping path $W$ is constructed as a set of associations with the time points of the two series repeating, it is not a warping function of the form $t_2 = f(t_1)$ that relates the two time scales. Moreover, the optimal warping path has a variable length $k$ and is not a regularly sampled time series.

Although the DTW algorithm locally matches the elements of the two sequences, the main goal is the computation of an overall measure that describes the global similarity of the two sequences. A global similarity score between two sequences has been defined since the early days of the algorithm as the total costs of the optimal warping path, i.e. the sum of the distances in the similarity matrix $M$ along the optimal warping path $W$:

$$D_{DTW}(S_1, S_2) = \sum_{i=0}^{k-1} M(w_i)$$

Equation 2

The above score emphasizes the differences in the amplitude of the sequences while it attempts to supress differences in phase. More recently, another global similarity measure was proposed by Folgado et al. [26]: the Time Alignment Measurement (TAM) quantifies similarity in the temporal domain. It utilizes the length $k$ of the warping path to measure the total amount of warping needed by decomposing it into the sum of its diagonal, horizontal and vertical segments $k = L_{diagonal} + L_{horizontal} + L_{vertical}$. The length of the diagonal segments reflects the total time in which the two sequences are in phase, while that of the horizontal and vertical segments reflect the amount of warping. TAM sums the lengths of the vertical and horizontal segments to measure the accumulated time warping through the entire length of the warping path. It furthers normalizes the result to create a universal scale with the definite range [0, 3] for any pair of sequences. TAM captures changes in the relative phase between two sequences rather than the relative phase itself. Sequences with the same shape would yield a low TAM even when they have a constant phase difference. In contrast, small variations of the warping path around the diagonal may result in large TAM values, for instance, as the result of overfitting random noise variation in the signals. TAM therefore captures shape "deformations" arising from time distortion. A current limitation of TAM is that it can only be computed for warping paths that include exclusively single forward steps.

The quality of the alignment achieved by DTW depends greatly on the distance metric used in the construction of the similarity matrix $M$. Often, the time series themselves may be processed prior to their alignment to facilitate the comparison of the relevant features. For instance, sequences may be z-normalized, rescaled or differentiated [10]. The point-to-point character of the distance matrix has also been criticized as creating singular points of non-realistic extreme time compression or stretching, and not being able to model the serial correlation of time points [14]. The recently proposed shapeDTW method [9] addresses these shortcomings of the simple Euclidean distance matrix, by using local shape descriptors that characterize the shape of the sequence in the neighbourhood of each time point. The distance matrix is then constructed from the differences between those descriptors instead of the data points of the sequences, resulting in a feature-to-feature instead of a point-to-point match. The simpler such descriptor is the raw-subsequence, which was proven to be effective in many cases [9]. A similar sliding window descriptor was proposed by Folgado et al. [26] to improve the accuracy of TAM and reduce overfitting.

Although global DTW based measurements and distances as the above are relatively straightforward to define and may effectively capture the overall similarity between sequences, at the local level, the processing of warping paths and the extraction of the local phase relations between the sequences is not trivial. Locally, when one sequence is compressed, the other is stretched by the same amount. The time warping is achieved by associating a single time point of one sequence with a segment that consists of several consecutive points of the other. While the magnitude of the phase difference is directly proportional to the length of the segment, neither of the individual time scales of the sequences is a suitable reference frame for locating the phase difference. A naïve approach is to create a time rescale function $t_2 = f(t_1)$, by arbitrary eliminating the repeating time points of $S_2$ in the warping path $W$. However, when creating the inverse rescaling function $t_1 = g(t_2)$, different time points, this time from $S_1$, would be eliminated, and therefore $g \neq f^{-1}$. In other words, measuring the phase difference by creating a time scaling function would depend on which sequence is chosen as the reference. This becomes especially important when dealing with a set of multiple sequences, where one of them must be arbitrarily chosen as the frame of reference for the entire set.

A notable application of the DTW algorithm is the averaging of time series. As in the case of sequence comparison, Euclidean averaging is sensitive to phase differences and variation, resulting in mean sequences with drastically different shapes than the collection they are supposed to summarize. To solve this problem and achieve a more meaningful average, DTW can be used to realign the sequences before averaging them.

Two general approaches have been followed. The first approach is based on successively averaging pairs of sequences in a certain order. The most representative methods following this approach are the Nonlinear Alignment and Averaging Filters (NLAAF) [29] and the Prioritized Shape Averaging (PSA) [16,17]. The NLAAF method successively reduces the number of sequences in a dataset by replacing each pair with its average, until there is only one pair left. PSA follows a hierarchical scheme, in which similar sequences are clustered together, and then averages are created for each cluster in a bottom-up process until the average of the entire dataset is computed, similarly to the agglomerative hierarchical clustering. Both methods average pairs of sequences by averaging their associated data points along the warping path. As the warping path is longer than any of the two sequences being averaged, the length of the mean sequence increases with each step. To prevent this, PSA uses uniform rescaling [16]. It assumes that the greater size of the mean sequence is due to a uniform lengthening, that is, of a finer and constant sampling interval. However, the warping path is not a regularly spaced time series. Therefore, this method introduces distortion in the temporal domain, as we illustrate in section 4. The distortion depends on the warping path and thus will be unique for each averaged pair. As more pairs are averaged, the distortions accumulate unpredictably resulting in poor phase and amplitude averaging.

The second approach, the DTW barycenter averaging (DBA) [3,15], attempts to fix the issue of the warping path length by using an initial sequence taken from the dataset to align all other sequences to it. The alignment is achieved by creating time warp functions for all sequences $p$ relative to that initial reference $t_p = f(t_{ref})$, similarly to the naïve method described above. Then, the alignment and mean sequence are iteratively refined. This method was shown to be superior to both PSA and NLAAF methods [3]. A main concern with this approach is that there is no clear criterion for choosing or synthesizing an initial sequence. Therefore, the method is not deterministic and the temporal characteristics, such as the position of features in time and their duration, are biased by the choice of the initial sequence, and therefore do not correspond to the average temporal characteristics of the dataset. While several adaptations of DBA have been proposed to reduce the impact of the initialization and improve the shape of the mean sequence produced by DBA [30–32], these methods are still prone to initialization biases and result in inexact phase averaging.

DBA and its derivatives aims at minimizing the Fréchet equation [20] commonly referred to as the Within Group Sum of Squares (WGSS) of the dataset with respect to the mean sequence $S_{av}$ and the DTW distance:

$$WGSS|_{DTW} = \sum_{p=1}^{N} D_{DTW}^{2}(S_{av}, S_p) \qquad \text{Equation 3}$$

where, $N$ is the number of sequences in the dataset. The WGSS has been used as a general criterion for comparing and evaluating different averaging techniques [3]. The principle is that in general an average object is found at the "centre" of the objects it averages and, therefore, must minimize the sum of the square of its distances to the objects in the dataset. If the mean sequence A produced by one method corresponds to a lower WGSS than the mean sequence B produced by a different method, then sequence A is closer to the theoretical centre. Different centres may be defined based on different distance metrics used in the computation of the WGSS.

The minimization of $WGSS|_{DTW}$ provides a sound theoretical definition of the mean object in the space of sequences in which the distance is determined by the DTW score of Equation 2, and is well suited for many tasks. However, the DTW distance aims at supressing phase differences between the sequences. During the computation of the $WGSS|_{DTW}$, the average sequence $S_{av}$ is realigned to each sequence $S_p$ before computing the respective distance. Even large temporal distortions and biases introduced in the averaging process would be supressed during the computation of the $WGSS|_{DTW}$. In fact, the authors in [3] propose an adaptive scaling technique to shorten the mean sequence by eliminating certain of its elements without increasing the $WGSS|_{DTW}$. Thus, a sequence minimizing the $WGSS|_{DTW}$ does not necessarily represent the centre of the phase variation in a dataset.

Therefore, current DTW methods of averaging time series do not accurately capture the centre of the phase variation between multiple sequences. They either suffer from non-uniform time distortion inherent in the construction of the warping path, or their definition of the mean object is not the centre of the phase variation but the centre of the variation of the DTW distance. This paper addresses the need for an accurate estimation of the phase variation and a corresponding mean sequence. In other words, we seek to develop a generic and automated method for the computation of a mean sequence that averages the phase and amplitude variation as independent dimensions in a Euclidean space.

## 3 Time Warp Profile

Before describing the averaging algorithm in section 4, we need to introduce a core concept of the method, the Time Warp Profile (TWP). The main aim of the TWP is to address the issue of the variable length of the warping path by transforming it into a time sequence of a constant sampling interval and a predetermined length. At the core, lies the conversion of the DTW steps, from steps in the two independent time scales of the sequences ($t_1$, $t_2$), to steps along the common diagonal of the similarity matrix $M$. From these steps, we can reconstruct the DTW path in the new frame of reference. We begin by presenting the frame of reference along the diagonal (section 3.1) and then the conversion of the basic DTW step pattern and the construction of the TWP (section 3.2). Based on the TWP, we propose, in section 3.3, a new distance measure that measures the overall phase difference between two sequences. Finally, we generalize the TWP and TAM for any DTW step pattern in sections 3.4 and 3.5, respectively.

### 3.1 The phase frame of reference

Let us consider the similarity matrix $M$ corresponding to two time sequences $S_1$ and $S_2$ (Figure 2, Left, black lines). Along the main diagonal the two sequences are in phase, that is, time advances equally for both. Deviations from that diagonal correspond to time warping, in which one sequence advances slower than the other. Below the main diagonal, $S_2$ lags behind $S_1$, and above the main diagonal, $S_1$ lags behind $S_2$. The larger the deviation from the main diagonal the larger the respective lag.

The coordinates of the time points of the similarity matrix $M$ can be expressed in a reference frame along its main diagonal, similar to the rotated frame of reference used by Zhou et al. [33] in the computation of the "optimal thermal causal path", a technique analogous to DTW. One dimension of the alternative frame of reference is the projection of a point ($t_1$, $t_2$) of the matrix $M$ on the main diagonal and corresponds to the time in which the two sequences are in phase. The other dimension is the distance from the main diagonal and indicates the amount of deviation from the in-phase alignment, in other words, the lag between $S_1$ and $S_2$. This reference frame may be constructed from the diagonals and anti-diagonals that traverse the similarity matrix (Figure 2, blue lines). If $n$ and $m$ are the lengths of $S_1$ and $S_2$ respectively, then there are $n+m-1$ diagonals and an equal number of anti-diagonals that form a lattice. We will refer to this lattice as the *phase frame of reference*.

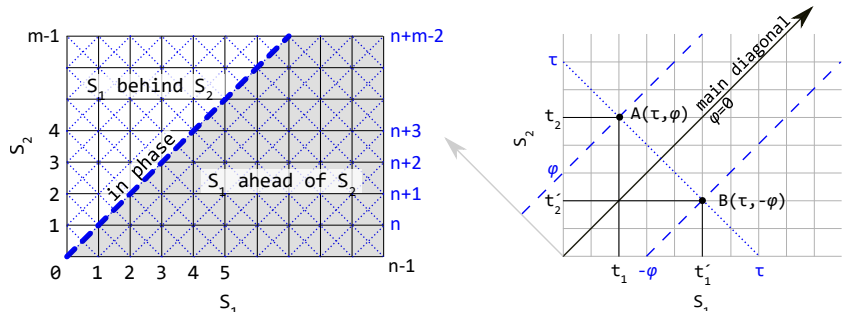

*Figure 2. LEFT: Similarity matrix as lattices. In black horizontal and vertical lines the lattice constructed from the time points of the two sequences $S_1$ and $S_2$. In blue diagonal dotted lines, the phase frame of reference constructed by the diagonals and antidiagonals. RIGHT: Coordinates in the phase frame of reference. The main diagonal has an index $\varphi = 0$. The diagonals above it have positive $\varphi$ indices and the ones below negative.*

The discrete indices of the diagonals and anti-diagonals may be used as coordinates for any element $(t_1, t_2)$ of the similarity matrix $M$, as shown in Figure 2 (Right). The in-phase time corresponds to the index $\tau$ of the antidiagonal that passes through that element of the matrix, and the deviation from the in-phase alignment corresponds to the index $\varphi$ of the respective diagonal. The conversion between the two reference frames is summarized in the following system of equations:

$$\left.\begin{array}{l} \tau = t_1 + t_2 \\ \varphi = t_2 - t_1 \end{array}\right\} \rightarrow \begin{array}{l} t_1 = (\tau - \varphi)/2 \\ t_2 = (\tau + \varphi)/2 \end{array} \qquad \text{Equation 4}$$

## 3.2 Transforming the warping path

The simple transformation proposed here expresses the warping path in the *phase frame* of reference. The advantage of the *phase frame* becomes apparent when we look at how the steps used to construct a warping path may be expressed as steps in the $(\tau, \varphi)$ dimensions (Figure 3 and Table 1). We observe that any DTW step results in an advancement along the main diagonal, that is in an increase of the in-phase time $\tau$. In other words, each value of $\tau$ corresponds to only one value of $\varphi$ and, therefore, the DTW path in the *phase frame* is expressed as a function $\varphi(\tau)$. This is a direct consequence of the monotonicity requirement in the construction of any warping path.

A second observation is that a step parallel to the diagonal (i.e. a step that increases equally indices $t_1$ and $t_2$) increases the value of $\tau$ by 2, without altering the value of $\varphi$. Therefore, the function $\varphi(\tau)$ is not uniformly sampled along time $\tau$. To correct this, we may break down the longer diagonal steps into two steps, each of which increases

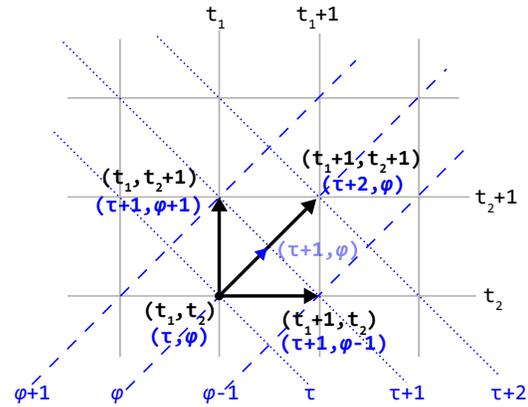

Figure 3. The three common DTW steps expressed in time sequences' indices $(t_1, t_2)$ and in the phase frame of reference $(\tau, \varphi)$. The single step $\Delta t = (1,1)$ is broken down to two steps in the phase frame.

|  | Indices $(\Delta t_1, \Delta t_2)$ | Phase Fame $(\Delta \tau, \Delta \varphi)$ |
|---|---|---|
| Steps | (1,1) | $2 \times (1,0)$ |
|  | (1,0) | $(1, -1)$ |
|  | (0,1) | $(1, +1)$ |

Table 1: The three common DTW steps expressed as changes in the time sequences indices and as changes in the diagonal and anti-diagonal indices of the phase frame

---

**Algorithm 1: Time Warp Profile (TWP)**

---

**Input**: DTW path $w_i = (t_1, t_2), i \in \{0, \ldots, k-1\}$

**TWP**:

1. Initialize variable $\varphi_0 = 0$ and index $\tau = 0$
2. Compute the DTW steps $\Delta w_i = w_{i+1} - w_i = (\Delta t_1, \Delta t_2)_i$
3. **For** all $\Delta w_i, i \in \{0, 1, \ldots, k-1\}$

    If $\Delta w_i = (1,1)$      $\varphi_{\tau+1} = \varphi_\tau \,;\, \varphi_{\tau+2} = \varphi_\tau \,;\, \tau = \tau + 2$
    else if $\Delta w_i = (1,0)$    $\varphi_{\tau+1} = \varphi_\tau - 1;\, \tau = \tau + 1$
    else if $\Delta w_i = (0,1)$    $\varphi_{\tau+1} = \varphi_\tau + 1;\, \tau = \tau + 1$

    **End For**

**Output**: array $\varphi_\tau$

---

the value of $\tau$ by 1, without introducing any time warping. In this way, we may evaluate the values of the function $\varphi(\tau)$ at all anti-diagonals $\tau$ and obtain a sequence $\varphi_\tau$ with $\tau \in \{0, \ldots, n+m-2\}$. Then, $\varphi_\tau$ has a fixed length $(m+n-1)$ that depends only on the sizes of sequences $(n, m)$ and not on the length $k$ of the warping path. In other words, the above transformation converts the DTW path into a time sequence of a finer but constant sampling rate that is double that of the initial sequences (Figure 4). It follows from Equation 4, that the values of $\varphi_\tau$ represent the local lags between the initial time sequences $S_1$ and $S_2$, and the index $\tau$ corresponds to the time in which the two sequences are in phase. The optimal alignment of the two sequences warps symmetrically their individual time scales $t_1$ and $t_2$ by $\mp \varphi_\tau/2$ relative to their *in-phase* time $\tau$. We call the time series $\varphi_\tau$ the *time warp profile* (TWP) of the pair $(S_1|S_2)$. To compute the TWP, we begin at the origin $\varphi_0 = 0$ and proceed applying the warping path steps to all subsequent $\varphi_\tau$ as shown in Algorithm 1 and Figure 4. As the DTW algorithm produces a symmetric warping path for a transposed similarity matrix, it follows from Equation 4 that $\varphi(S_2|S_1) = -\varphi(S_1|S_2)$.

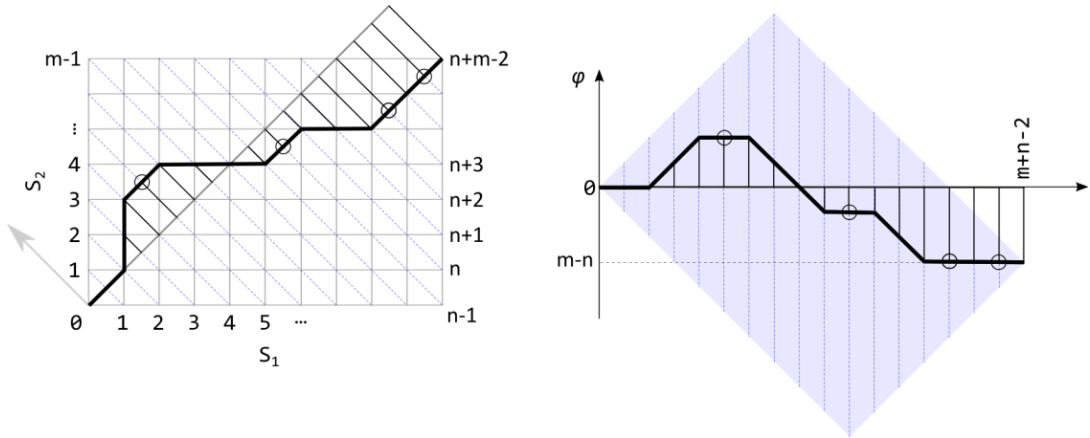

*Figure 4. An example of a DTW path as associations in the similarity matrix (left) and converted into a TWP in the phase frame (right). The points marked with circles correspond to points in the phase frame that fall in between the time points of the initial sequences.*

Converting the warping path into a time sequence does not change the warping path as it neither introduces new nor omits existing information. Breaking down a diagonal step into two steps occurs only for DTW steps that are parallel to the diagonal, that is, for steps in which the phase between the two sequences is preserved. Although this results in elements of $\varphi_\tau$ that fall in between elements of the similarity matrix (Figure 4, circled points), these elements do not introduce any time warping and no element of the warping path is omitted.

### 3.3   Phase distance

In principle, it is possible to compute a phase distance directly form the warping path $W$ and define it as the sum of the square of the difference between the associated time points $w_i$: $\sqrt{\sum_{i=1}^{k}(t_1(i) - t_2(i))^2}$. However, such a measure would depend on the length $k$ of warping path and would weigh the phase differences in the vertical and horizontal segments of the path disproportionally relative to the diagonal ones. This may be crucial considering the potential overfitting character of DTW and the fact that similar phase differences between signals may correspond to drastically different numbers of diagonal elements. Instead, we propose a measure of domain or phase difference, the *Phase distance* $\Phi(S_1, S_2)$, that is based on TWP (Equation 5). As TWP has a fixed and predetermined length, the *Phase distance* is robust against overfitting. It may be thought of as the Euclidean distance between sequences $S_1$ and $S_2$ in the time domain, although it lacks the metric property.

$$\Phi(S_1, S_2) = \sqrt{\sum_{\tau=0}^{n+m-2} \varphi_\tau^2(S_1|S_2)} \qquad \text{Equation 5}$$

### 3.4   Generalization of the Time Warp Profile

So far, we derived the TWP (Figure 4) for DTW paths produced from the basic step pattern consisting of single horizontal, vertical, and diagonal steps (Table 1). To generalize TWP to any warping step pattern we present an alternative method for computing it. Using Equation 4 to convert all points of the DTW path to the *phase frame of reference* we obtain two coupled arrays $(\tau, \varphi)$ with length $k \leq m + n - 1$ and values of $\tau$ that are not necessarily consecutive. As $\varphi$ is always a function of $\tau$, and therefore all $\tau$ values are unique, $\varphi$ can be interpolated at the missing $\tau$ values to obtain the time sequence $\varphi_\tau$. Algorithm 2 computes the TWP for warping paths that include any

Sioros et al.                                                                    12

---

**Algorithm 2: Generalized computation of TWP**

---

**Input**: DTW path $w_i = (t_1, t_2), i \in \{0, \ldots, k-1\}$

**TWP**:

1. Compute $(\tau, \varphi)$ of length $k$ from Equation 4.
2. Linear interpolation of $\varphi$ for $\tau \in \{0, \ldots, n+m-2\}$ to obtain $\varphi_\tau$.

**Output**: array $\varphi_\tau$

---

alternative step pattern. Then, $\Delta\varphi_\tau$ can take any non-integer value in the range $[-1, 1]$. For the basic DTW step pattern, the two algorithms are equivalent.

### 3.5 Generalization of the Time Alignment Measure

In this subsection, we present an application of TWP in the generalization of TAM. As we saw in section 2, TAM can only be computed for warping paths consisting of the basic DTW steps. Here we generalize TAM to any warping path. Using as point of departure the original TAM equations [26], we express TAM as a function of the partial derivative $\Delta\varphi_\tau$ of TWP with respect to $\tau$, which can be computed for any warping path irrelevant of the step pattern used for its construction.

We first examine the case where the two sequences have equal lengths *n=m*. Let $\Delta\varphi_\tau$ be the derivative of the TWP in the *phase frame*, $\Delta\varphi_\tau = \varphi_{\tau+1} - \varphi_\tau$, which for the common single step DTW pattern $\Delta\varphi_\tau \in \{-1,0,1\}$. Let $\vec{\delta}$, $\overleftarrow{\delta}$ and $\bar{\delta}$ be the vertical, horizontal and diagonal steps in the warping path [26 eq. 11]. It follows that:

$$\sum_{i=1}^{k} \vec{\delta}_i + \sum_{i=1}^{k} \overleftarrow{\delta}_i = \sum_{\tau=0}^{2n-3} |\Delta\varphi_\tau| \qquad \text{Equation 6}$$

$$2\sum_{i=1}^{k} \bar{\delta}_i = 2n - 2 - \sum_{\tau=0}^{2n-3} |\Delta\varphi_\tau|$$

The TAM distance [26 eq. 12][1] can then be written in terms of $\Delta\varphi_\tau$ as:

$$\Gamma = \frac{3}{2(n-1)} \sum_{\tau=0}^{2n-3} |\Delta\varphi_\tau| \qquad \text{Equation 7}$$

---

[1] In [26] there is a conflict between eq. 12 and the definition of $\Gamma \in [0,3]$. As the advance, delay and in phase elements are only defined based on the steps of the DTW path (eq. 11), the maximum number of elements for $\vec{\delta}_k$, $\overleftarrow{\delta}_k$ and $\bar{\delta}_k$ is n-1, m-1 and min(n,m)-1, which should also correspond the normalization factors in eq 12. Here, we use these corrected factors.

Equation 7 is equivalent to equation 14 in [26]. In a similar fashion, it can be shown that the relation between TAM and the derivative $\Delta\varphi_\tau$ for sequences of non-equal lengths can be written in terms of the lengths of the sequences $L = max(m,n) - 1$ and $\ell = min(m,n) - 1$ as:

$$\Gamma = \frac{1}{2L\ell}\left((2L+\ell)\cdot\sum_{\tau=0}^{L+\ell-1}|\Delta\varphi_\tau| - (2L-\ell)\cdot|\varphi_f|\right) \qquad \text{Equation 8}$$

For $m = n$, the second term is eliminated and the first leads to the simpler Equation 7. Equation 8 can serve as the generalized TAM for any DTW step pattern, in which case $|\Delta\varphi_\tau| \leq 1$. It is trivial to show that the properties of this generalized TAM are holding; for example, it is still bound by the same upper and lower limits $\Gamma \in [0,3]$.

When $m \neq n$ the minimum value of $\Gamma$ is: $\Gamma_{min} = \frac{|\varphi_f|}{L} = \frac{L-\ell}{L}$ which corresponds to the overall time stretching required to meet the global constraint of DTW. The remainder $\Gamma_w = \frac{2L+\ell}{2L\ell}\left(\sum_{\tau=0}^{L+\ell-1}|\Delta\varphi_\tau| - |\varphi_f|\right)$ represents the accumulation of time warping due to local phase differences between the signals. Thus, TAM can be written as the sum of these two terms $\Gamma = \Gamma_{min} + \Gamma_w$.

## 4 Averaging phase variation in time series

Let us consider the 2-dimensional space, in which one dimension is amplitude (vertical) and the other is time (horizontal). We seek to produce a mean sequence that simultaneously averages the characteristics of a set of sequences in both dimensions, that is, to produce a Euclidean average both amplitude- and domain-wise. The aim of our method is to use the "horizontal" associations between the time points of the sequences of the dataset, to disentangle the phase from the amplitude variation and produce a sequence that accurately averages the phase variation. The problem then can be formalized as the minimization of the WGSS with respect to the Phase distance $\Phi$, instead of the DTW score:

$$WGSS|_\Phi = \sum_{i=p}^{N}\Phi^2(S_{av}, S_p) \qquad \text{Equation 9}$$

### 4.1 Averaging two time sequences

We begin the description of our method with the computation of the mean of two sequences. As we saw in section 3, the time scale $\tau$ represents the in-phase time of the pair. From Equation 4, we compute the aligned versions of the two sequences along the time scale $\tau$, which are then averaged:

Sioros et al.                                                                                                                                                           14Sioros et al.                                                                                                                                14

$$\left.\begin{array}{l}S_1(\tau) = S_1\left(\left\lceil\left|\frac{\tau - \varphi_\tau}{2}\right|\right\rceil\right)\\S_2(\tau) = S_2\left(\left\lceil\left|\frac{\tau + \varphi_\tau}{2}\right|\right\rceil\right)\end{array}\right\} \to S_{1|2}(\tau) = \frac{S_1(\tau) + S_2(\tau)}{2} \qquad \text{Equation 10}$$

where $S_{1|2}(\tau)$ is the mean sequence of the pair (Figure 5, bold red line). The TWP has a fixed length independently of the length of the warping path, so that the aligned counterparts of the two sequences also have the same fixed length that depends only on the lengths $L_1$ and $L_2$ of the initial sequences. As the sampling rate of $\tau$ is double of the initial sampling rate, we can compute the mean $S_{1|2}(\tau)$ for every other value of $\tau$ producing a mean sequence with length $L = \lceil(L_1 + L_2 - 1)/2\rceil$.

From Equation 4, it follows that the time scale $\tau$ is equidistant from the two individual time scales. In this sense, $\tau$ is the average of the individual time scales and, therefore, by definition, minimizes the Phase distance from them. The amplitude difference between the two aligned sequences is also minimal by definition of the optimal warping path. Taking the point-to-point arithmetic mean between the two aligned sequences produces an accurate average in both the time and amplitude dimensions simultaneously. The features that are aligned by the DTW algorithm are preserved and their "horizontal" phase is averaged. This is reflected in the accurate estimation of the location of the two landmarks A and B, compared to their theoretical arithmetic mean.

Figure 5 shows additionally the average sequence produced from the same warping path according to the PSA method, i.e. first, by calculating the arithmetic mean of the associated time points along the warping path, and then using uniform scaling to produce a sequence with the correct length [16]. As it is mentioned in section 2, this method introduces distortion which is evident in the figure since the location of the two landmarks deviates from the expected theoretical mean. The distortion comes about from the fact that the associations in the warping part do not form a regularly sampled sequence. As we can see from the construction of the TWP, while the non-warped segments are sampled at the same sampling interval as the initial sequences, the warped segments are sampled at twice that rate. The different parts are not distinguished when the mean sequence is uniformly scaled to compensate for its greater length. Therefore, the non-time warped segments are shortened in duration while the time warped segments are stretched. In the example of Figure 5, landmark A appears earlier than theoretically expected as the initial flat part of the signals is not time warped. In contrast, the mean curve appears to be stretched, having a longer duration than expected, for the most extremely warped segment (inset of Figure 5). In general, the amount and shape of distortion will depend on the particularities of the warping path.

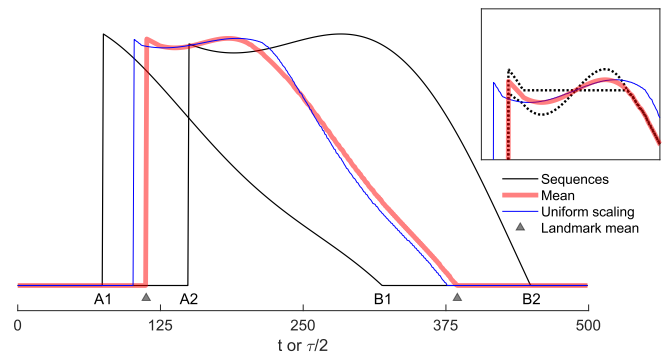

*Figure 5:Averaging a pair of sequences. A and B denote two landmarks and the symbols (▲) denote their respective theoretical mean location. The thin blue line corresponds to the average produced by the uniform scaling applied in PSA [16]. The inset shows a part of the signals where the most extreme warping is applied. The dotted black lines in the inset correspond to the aligned versions of the two sequences along τ.*

Alternatively to the PSA method, it has been previously suggested that the averaging of two associated time points should not only average the data points but also their time coordinates [18,34], which essentially locates that average along the time scale $\tau$. However, as these average time points are not equally spaced, the resulting time series has no constant sampling rate, requiring resampling and interpolation, which in turn leads to information loss and degradation of the signals. In contrast, averaging along the TWP produces no distortion, because TWP has, by definition, a constant sampling rate, while at the same time the averaged elements are all time points associated by the optimal warping path and therefore require no further interpolation.

### 4.2  Averaging multiple sequences

We apply the above principle of pairwise averaging in approximating an accurate domain-wise mean sequence of a dataset. The method follows the iterative process described in detail by Algorithm 3. Here, we present the general principle behind it. In each iteration, a pair of sequences in the dataset is averaged according to Equation 10 and they are replaced in the dataset by their mean. While the two sequences may initially differ in phase, shape, or both, replacing them with their mean averages out those differences. As more and more pairs are averaged, the sequences in the dataset gradually become more and more similar, with the common features in the dataset being aligned and other differences being smoothed out. To ensure fast convergence, each iteration averages the most dissimilar pair of sequences in the updated dataset. After several iterations, the inter-sequence distances are reduced and the maximum distance falls below a predetermined threshold, which leads the process to termination. Finally, the mean sequence is computed as a simple point-to-point arithmetic mean of all sequences in the updated dataset.

The distance used to determine the most dissimilar pair at each iteration is the Euclidean distance. For sequences of non-equal lengths $n$ and $m$, we propose the following simple scaling:

$$DS_{euc}(p,q) = \frac{n}{m} \sum_{i=0}^{m-1} \left(S_p(i) - S_q(i)\right)^2, \text{with } n > m \qquad \text{Equation 11}$$

The novelty of the above method lies in two central features. First, averaging a pair of sequences along the TWP corrects the time distortion produced in previous methods due to the variable length of the warping paths. Second, our method does not use pairwise averages to compute means of progressively bigger subsets of the initial dataset, as in PSA or NLAAF. Instead, the pairwise means aim at averaging out the phase variation within the dataset by iteratively realigning the sequences. As the final average is only computed at the end from the realigned sequences, the ordering effect is minimized. Furthermore, in contrast to DBA, our method realigns the sequences to their "average time scale" and not to an arbitrary initial sequence. The time scales of the pairwise mean sequences gradually converge towards a global average time scale, while the length of the final mean sequence converges towards the mean length of the initial sequences. This process, on the one hand, corrects the biases and distortion arising from the choice of initialization, and on the other hand, ensures a deterministic solution.

Finally, the standard deviation $\sigma(t)$ of the phase distribution around the mean sequence is computed as a function of the time scale of the mean sequence. First, the optimal warping paths $W_p$ of each sequence $S_p$ with $S_{av}$ are determined through DTW. For each warping path, a time warping function $t_p = f_p(t_{av})$ is computed, where $t_p$ and $t_{av}$ represent the time points of $S_p$ and $S_{av}$ respectively. The function $f$ is computed in the

---

**Algorithm 3: TWP average**

**Input**: Dataset of $N$ sequences $S$; threshold $Thr$

**Average:**

  Compute Distance Matrix $D_{pq} = DS_{euc}(S_p, S_q), \forall p, q \in \{1, \dots, N\}$ (Equation 11)

  **While** $max(D) > Thr$

  1. Find pair of sequences $(S_p, S_q)$ where $D_{pq} = max(D)$
  2. Compute $\varphi_\tau(S_p|S_q)$ from Algorithm 2
  3. Compute pairwise mean sequence $S_{p|q}$ from Equation 10
  4. Replace sequences p and q with mean: $S_p \leftarrow S_{p|q}, S_p \leftarrow S_{p|q}$
  5. Update elements in distance matrix:

  $D_{pi} = D_{qi} = DS_{euc}(S_{p|q}, S_i), \forall i \in \{1, \dots, N\}$
  **End while**

  Compute $S_{av} = \frac{\sum_{i=1}^{N} S_i}{N}$

**Output**: $S_{av}$

naïve way described in section 2, that is, by eliminating the repeating time points $t_{av}$, keeping only the first of each repeated point in the warping path. The standard deviation is then computed as:

$$\sigma(t_{av}) = \sqrt{\frac{\sum_{p=1}^{N}(f_p(t_{av}) - t_{av})^2}{N}}\qquad\text{Equation 12}$$

### 4.3 Computational complexity

Here, we calculate the computational complexity of the TWP averaging method and compare it to that of DBA, one of the most efficient methods to date. Each iteration of the TWP averaging algorithm consists of: 1) the computation of the mean of a pair of sequences (steps 2 and 3 in Algorithm 3), and 2) the update of the respective elements in the pairwise distance matrix $D$ (steps 4 and 5). Step 2 involves the computation of the DTW between two sequences and step 3 the averaging along the TWP. Their combined complexity is $\Theta(DTW) + \Theta(mean) = \Theta(L^2 + 3L)$. Steps 4 and 5 involve the computation of the Euclidean distance between the rest of the sequences in the dataset and the newly computed pairwise mean, which requires $(N-2) \cdot L$ operations. If we designate $n$ the number of iterations, then the total complexity of the iterative process is $\Theta(n(L^2 + 3L + (N-2)L))$. If we express the total number of iterations as a factor of the total number of sequences $n = f \cdot N$, then the complexity of the iterative process may be written as $\Theta(f(NL^2 + N^2L + NL))$.

Algorithm 3 includes two more computations which are not repeated: the computation of the initial distance matrix between all sequences in the dataset, which requires $(N^2L)$ operations, and the computation of the final overall mean sequence from the updated dataset, which requires $(NL)$ operations. The overall complexity of our algorithm is therefore the combination of the complexity of the iterative process and the non-iterative steps: $\Theta(fNL^2 + (f+1)N^2L + (f+1)NL)$. The expression may be simplified by dropping the lower order term that contributes significantly less than the second order terms: $\Theta(fNL^2 + (f+1)N^2L)$.

DBA requires $(I \cdot (NL^2 + NL))$ operations [3], where $I$ is the number of DBA iterations typically set to $I=15$. As previously, it may be simplified to $\Theta(INL^2)$. In most practical applications TWP convergences faster than the required iterations by DBA ($f < I$), so that, the efficiency of DBA is greater than that of TWP only for large datasets. In general, the TWP method is more efficient than DBA under the following condition:

$$f < \frac{I \cdot L - N}{L + N}\qquad\text{Equation 13}$$

As we present in section 5.2.2, the convergence of the TWP method meets the above efficiency threshold for the majority of the 20 datasets from the UCR archive [35] that we tested.

## 5 Experimental Evaluation

Real-world datasets commonly lack ground truth of their continuous phase variation so that a theoretical mean sequence is difficult to compute. For this reason, we first evaluate and demonstrate the TWP averaging method with a dataset of synthetic sequences (section 5.1), which allowed us to theoretically compute their mean. We then evaluate the method with real world data in section 5.2. First (section 5.2.1), we demonstrate how the mean sequence produced by our method may be used to retrieve information about the average characteristics of collections of signals. As illustrative cases, we constructed three datasets from the MAREA gait database [36]. We chose the MAREA database for two main reasons: 1) it contains sequences of accelerometer data and independently detected and synchronized events that we use to compute ground truth mean values for the phase variation of specific landmarks, and 2) the datasets extracted from the MAREA database contain extensive phase and amplitude variation, since they are collected from different participants in a variety of walking conditions. Both in the synthetic time series and the MAREA datasets, besides the evaluation against the ground truth, we also compare the TWP method against DBA and PSA. Finally, in section 5.2.2, we compare our method to DBA in terms of WGSS with respect to three different distance measures. To this end, we use 20 standard datasets from the UCR clustering and classification archive [35]. The standard single-step pattern was used in the construction of all optimal warping paths and all DTW computations in this section. In the figures of this entire section, the depicted sequences were vertically translated to allow for easier comparison between mean sequences produced by the different methods. For this reason, the vertical axes were omitted.

### 5.1 Synthetic time series

The synthetic sequences are all constructed from the same basic shape with randomized locations and amplitudes of the landmarks shown in Figure 6. The landmarks are randomized independently of one another so that not only the position but also the duration and relative amplitude of the various features vary across the dataset. Finally, white noise is added as random numbers. We produced in total 50 pseudorandom sequences of 500 points each.

A theoretical mean sequence is produced from the same basic shape as the pseudorandom sequences using the mean location and amplitude of each landmark. As we are aiming at averaging phase variation, the same level of white noise, which does not depend on phase, was added. The standard deviation of the temporal location of the six landmarks is also computed.

Mean sequences were computed using the TWP, DBA and PSA methods. For the computation of the TWP mean, the threshold $Thr$ for the termination condition was set to 0.0005 of the mean of the sum of squares of the initial sequences across the dataset (MSS). The MSS may be considered representative of the scale of the sequences and therefore it serves as the normalization factor for the threshold *Thr*. In addition, the standard deviation from the TWP mean was computed from Equation 12. The DBA algorithm was randomly initialized and run *I*=15 iterations. The dataset and all computed mean sequences are shown in Figure 7.

A visual inspection shows that the TWP mean sequence accurately reproduces the theoretical mean (GT), with the random manipulation of both amplitude and phase averaged out. All landmarks and their location fall within one sample from the mean location computed from their theoretical distribution. In contrast, neither DBA nor PSA successfully reproduce any of the landmark locations.

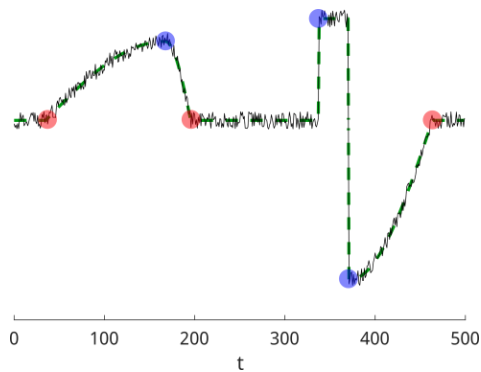

*Figure 6. Example of construction of a sequence for the synthetic dataset. The final sequence (solid noisy line) is constructed from a basic shape (dashed green line) consisting of 6 landmarks marked with blue and red circles. The temporal location of all 6 landmarks is randomized. In addition, the amplitude of the 3 blue landmarks is also randomized. White noise is added as random numbers in the range of ±10% of the maximum height.*

To quantitatively compare the TWP, DBA and PSA mean sequences, we compute their Euclidean, Phase and TAM distances from the theoretical mean (Table 2). The distance between two sequences containing similar levels of white noise will partially depend on the random structure of that noise and how it is aligned. Therefore, distances below a certain threshold would reflect differences only in the structure of the white noise and should be considered negligible. We defined *noise Thresholds* for the three distance measures (Euclidean, Phase and TAM) as the distance between two sequences that differ only in the structure of the white noise. To average out any effect of a particular instance of noise, Table 2 reports average values across 500 theoretical mean sequences that contain different randomly generated noise but are otherwise identical.

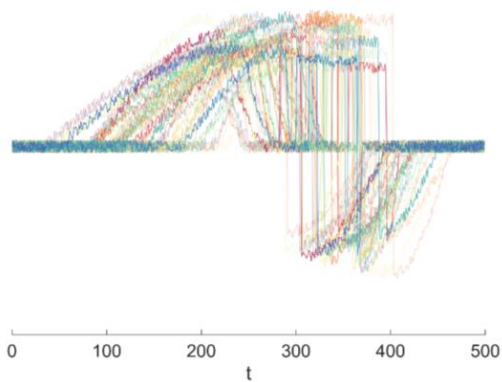
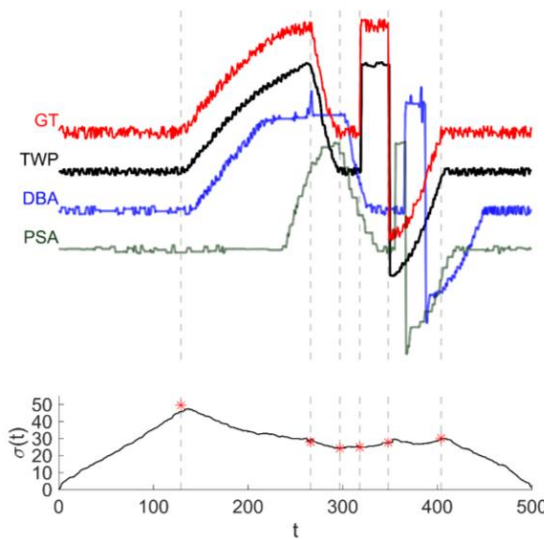

*Figure 7. Averaging a dataset of 50 synthetic time series. LEFT: the 50 pseudo-random time sequences. RIGHT Top: The theoretical ground truth (GT), TWP, DBA and PSA mean sequences. The dashed vertical lines correspond to the mean location of the landmarks. BOTTOM: the standard deviation from the TWP mean computed from Equation 12 as a function of time. The red \* correspond to the theoretical standard deviation of the landmarks.*

The quantitative evaluation shows that the TWP mean sequence is significantly closer to the theoretical mean than DBA or PSA with respect to any of the three distance measures. DBA and PSA perform similarly, with DBA slightly better in terms of the Phase measure. In comparison to the noise thresholds, the TWP mean achieves a below threshold TAM, which shows that it accurately reproduces the theoretical shape with no significant time distortion. Furthermore, its Phase distance is only marginally above the noise threshold, with its ratio to the threshold being 1.08, while the respective ratios for DBA and PSA are 5.01 and 9.76. Finally, while the Euclidean distance is larger than the respective noise threshold, this is predominantly due to the one sample phase difference between the TWP and the theoretical mean at the square pulse part of the signal.

In Table 3, we compare the standard deviation of the temporal location of the 6 landmarks computed for the TWP mean from Equation 12 (Figure 7, bottom) and the respective theoretical values (red * in the same figure). Their maximum difference is 3.7 samples, while, for 4 out of the 6 landmarks, our method produces the expected theoretical result with less than one sample difference.

In Figure 8, we present the maximum normalized pairwise distance between the sequences at each iteration as a fraction of the MSS, which shows that the convergence of the averaging process is not smooth. In fact, each step of the process does not guarantee a decrease in the maximum pairwise distance as can be seen from the spikes on the curve. This is however expected because the alignment of each pair of sequences along $\tau$ may create unpredictable differences from the rest of the sequences in the dataset. Nevertheless, in general, an overall convergence is achieved after a small number of iterations relative to the size $N$ of the dataset.

| Distance (noise threshold) | TWP | DBA | PSA |
|---|---|---|---|
| Euclidean (0.91) | **2.03** | 9.86 | 9.92 |
| Phase (157.5) | **169.9** | 789.8 | 1537 |
| TAM (0.818) | **0.753*** | 1.161 | 1.209 |

Table 2. Euclidean, Phase and TAM distances of the three computed mean sequences from the theoretical mean. The lowest distance is shown in bold. The noise threshold is shown in parentheses. Distances below the noise threshold are marked with (*).

| TWP | 45.8 | 29.2 | 24.0 | 24.9 | 27.5 | 29.6 |
|---|---|---|---|---|---|---|
| Theoretical | 49.5 | 28.0 | 24.2 | 24.9 | 27.7 | 30.2 |

Table 3. Standard deviation of the phase variation at the location of the 6 landmarks computed from the TWP mean sequence and from the theoretical distribution.

In Figure 9, we present the mean sequence at different stages of the iterative process. It shows no significant improvement of the features of the mean sequence after the threshold of 0.05 for the maximum pairwise difference is reached. This threshold is reached after 101 iterations, roughly double the size of the dataset. The process first aligns the prominent features while other details are averaged out. Once this initial convergence is reached (threshold 0.05) the differences between the sequences are only arising from the non-yet aligned white noise. As the process continues, the DTW algorithm aligns more details and the white noise added in the signal gradually emerges. The threshold of convergence for a certain dataset may be set by inspecting the level of noise in the dataset and determining the acceptable tolerance for the Euclidean distance between sequences relative to that noise. As the aligned sequences are averaged in the last step of the algorithm, the noise beyond this threshold will be averaged out, as in a standard point-to-point arithmetic average, producing a realistic mean sequence in both time and amplitude.

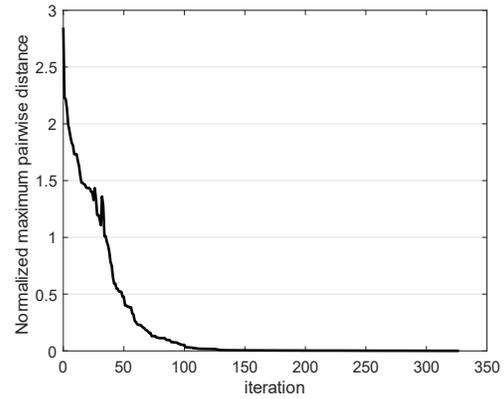

*Figure 8. Convergence of the TWP averaging algorithm*

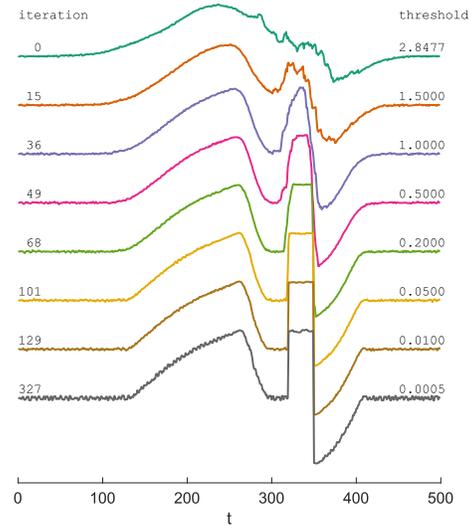

*Figure 9. The mean sequence at different iterations. Iteration 0 corresponds to the initial point-to-point arithmetic mean*

## 5.2 Evaluation with real world datasets

### 5.2.1 MAREA gait database

The MAREA dataset [36] contains time series collected from accelerometers (at 128Hz sampling rate) placed on the ankles of 20 participants and Force Sensitive Resistors (FSR) placed on their heels and toes while they were walking under different conditions. We used single axis data from the right foot, on which the accelerometer was randomly oriented and therefore have greater variability. The FSRs provide "heel-strike" and "toe-off" events that are synchronized to the accelerometer data. 11 subjects participated in indoor walking tasks that include walking on a treadmill at variable slopes and walking on a flat surface. 9 subjects participated in the outdoor task of walking on a pedestrian street.

We split the recordings for each task and subject into windows of 2.34s (300 time points) which correspond to a duration of at least 1 gait cycle. All windows included at least 2 "heel-strikes" to ensure that an entire gait cycle was captured in each window and exclude windows containing two half-cycles. The relative phase of the gait in each window was random and no specific alignment across windows was applied. This process resulted in 1178 windows for the "Treadmill", 265 windows for the "Indoor walk" and 288 windows for the "Outdoor walk" tasks.

Figure 10 shows 4 representative sequences from each of the three tasks together with the respective events from the FSRs. The relative amplitude and shape of the signals vary because of the random orientation of the accelerometer axes. The three tasks were conducted at a self-selected pace and included walking on different inclinations and surfaces. The process of randomly splitting the original time series into single gait cycles resulted in additional phase variation within each dataset. Therefore, the datasets contain significant variation in amplitude, shape, and phase.

Mean sequences were computed for each task separately, using the TWP, DBA, and PSA methods and a simple point-to-point arithmetic mean. The arithmetic mean is included to demonstrate the high variation of the sequences in each dataset. The termination threshold for the TWP algorithm was set to *Thr=0.01* of the initial MSS. The TWP averaging process reached its terminating condition after 6594, 1569 and 1589 iterations for the treadmill, indoor walk and outdoor walk sets, respectively. The iterations correspond to an *f* ratio over the number of sequences in each dataset of 5.6, 5.9 and 5.5, and therefore, according to Equation 13, TWP was more efficient than DBA in 2 out of the 3 datasets (the indoor and outdoor walk). The temporal location of the FSR events, "heel-strikes" and "toe-offs", were averaged, independently of the sequences in the dataset and of each other.

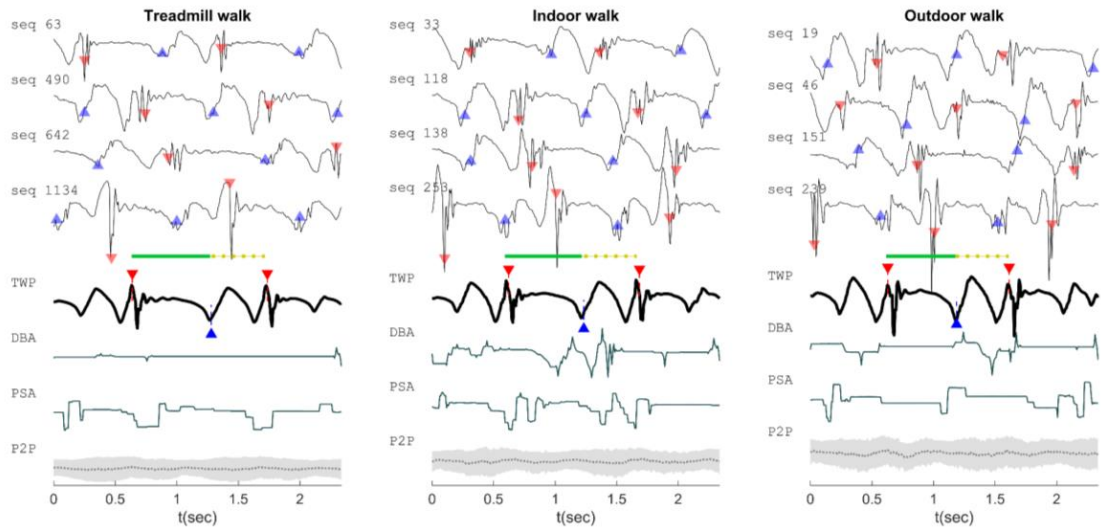

*Figure 10. The MAREA gait datasets and mean sequences. Each panel corresponds to a task/dataset. The top four curves are examples of the sequences in each dataset. Below them are the TWP, DBA, PSA and the point-to-point mean (P2P). The grey area around P2P corresponds to the standard deviation. The red ▼ represent the "heel-strikes" according to the FSRs, and the blue ▲ represent toe-offs. Their mean locations are marked on the TWP mean. The horizontal bars denote the stance (solid, green) and swing (dashed, yellow) phase of the gait cycle.*

Figure 10 presents the mean sequences below the example sequences of each dataset. The ground truth mean position of FSRs is depicted on the TWP mean sequence for comparison. The TWP mean sequence exhibits a typical gait cycle with two parts or phases: 1) the stance phase, the relatively flat part in which the foot is essentially idle, and 2) the swing phase, the part in which the foot moves forward. As the heel strikes the ground, the short fast oscillation of the acceleration arising from the impact may be observed. As the shock is rapidly attenuated, the oscillation is dampened. The mean location of the FSR events, which can be seen on the figure with coloured symbols (▼ for heel-strikes and ▲ for toe-off), correspond well to the two gait phases of the TWP mean sequence. Table 4 presents a quantitative comparison of the duration of the stance phase calculated as the duration between 1) the mean location of the toe-off (▲) and the first heel-strike event (▼), and 2) the peak and valley of the TWP mean sequence as marked with the solid horizontal bar in Figure 10. The durations of the entire gait cycle were calculated either from the two heel-strikes (ground truth) or from the corresponding peaks on the mean sequence. The maximum difference between the ground truth and the values estimated from the TWP mean is observed for the Treadmill walk (8ms) and it roughly corresponds to the duration of 1 time point (7.8ms).

The above results show that the TWP average offers a means to accurately estimate gait characteristics, such as mean durations of the different gait phases, from the accelerometer data alone. Alternative methods of estimation of gait properties would be based on the consistent characterization of the individual acceleration curves across participants, for instance through the automatic detection of landmarks, which is not trivial, especially considering the accelerometer axis was randomly oriented. In comparison, the mean sequences produced by the DBA or PSA algorithm fail to reproduce a consistent gait cycle shape and therefore it is not possible to estimate the duration of its different phases. Other signal features, such as the shock attenuation, may also be measured although they are not included in the current analysis as no relevant ground truth was available.

|     | Treadmill walk | Indoor walk | Outdoor walk |
|-----|----------------|-------------|--------------|
| FSR | 641ms (58.5%)  | 610ms (57.4%) | 558ms (56.6%) |
| TWP | 633ms (57.9%)  | 609ms (57.8%) | 555ms (56.4%) |

*Table 4. Stance phase duration in ms and as a percentage of the entire gait cycle duration.*

*5.2.2 UCR archive*

To quantitively compare the quality of the mean sequence produced by our method and DBA [3] we computed the ratios between the WGSS of the mean sequences produced by the two methods for 20 datasets from the UCR Time Series archive [35]. As this is not a classification task, the training and test datasets were joined together. First, each sequence was z-normalized. Then, we computed mean sequences for each class in a dataset with the TWP method (termination threshold *Thr*=10% of the initial MSS*)* and the DBA method (*I*=15). The global WGSS was then computed for the entire dataset using the respective mean sequence for each class and for three different distance measures: 1) the Phase distance *Φ* from Equation 9, 2) TAM substituting *Φ* with *Γ* in the same sum, and 3) the DTW distance from Equation 3. The Phase distance measures the absolute phase differences, and therefore the corresponding WGSS$|_\Phi$ reflects how well the mean sequence approximates the centre of the distribution of sequences in the time domain. TAM is sensitive to time distortion so that the corresponding WGSS$|_{TAM}$ reflects the quality of the shape of the mean sequence. Finally, we included the WGSS$|_{DTW}$ for comparison with previous studies, although it does not accurately assess phase variation. As the DBA method is not deterministic, we produced 12 different mean sequences for each class and dataset, corresponding to 12 random initializations of DBA. The ratios between the WGSS shown in Figure 11 offer a quantitative comparison of the mean sequences produced by the TWP and DBA methods. Ratios below 1 indicate that the WGSS of the TWP mean is less than that of DBA and therefore the TWP mean approximates better the "centre" of the sequences in terms of the respective distance measure. The variability in the ratios for each dataset reflects the effect of the random initialization of the DBA.

In all datasets and for all 12 DBA initializations, the TWP average outperforms DBA in the Phase and TAM based WGSS. Therefore, the TWP method provides a more accurate average of the phase variation and a more "distortion-free" mean sequence. The overall geometric mean ratio for the WGSS$|_{DTW}$ is close to 1, and the geometric mean ratios for each dataset split them equally between those below 1 and those above 1. However, we must note that many of the ratios above 1 are due to the poor alignment of DTW during the averaging process. For instance, in the Traces dataset, the alignment may be improved by rescaling the sequences in the range [0, 1] instead of z-normalizing them, which inverts the respective WGSS$|_{DTW}$ ratio in favour of the TWP method.

In terms of efficiency, we report that in 148 out of 178 classes from all datasets, the TWP algorithm converged faster than the efficiency threshold of Equation 13. In 13 out of the 20 datasets, TWP was more efficient in all classes, while the opposite was true for 5 datasets. Therefore, while the efficiency of the two algorithms depends on the dataset, in practice, TWP often outperforms DBA.

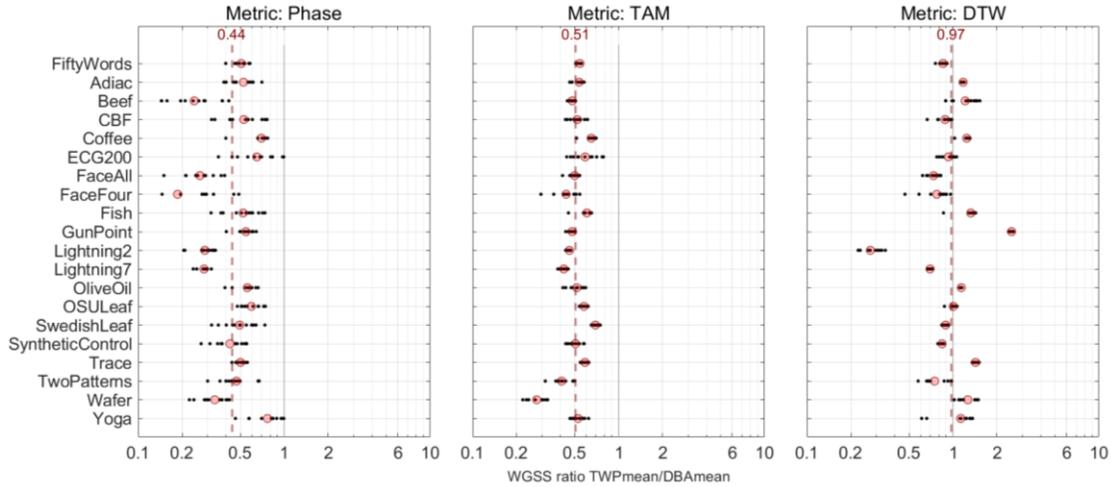

*Figure 11. Ratio of the WGSS computed for the TWP mean sequence over that for DBA for 20 datasets from the UCR archive [35]. WGSS based on three different distance measures are shown. LEFT: The Phase distance introduced in Equation 5. CENTRE: TAM distance. RIGHT: DTW distance. Black points on the graph represent the ratios for 12 random initializations of the DBA algorithm. Red circles represent the geometric mean of the 12 ratios for each dataset. The vertical red dashed lines represent the overall geometric mean across all 20 datasets which is also indicated by the number at the top of each graph.*

# 6 Conclusion

The averaging of multiple time series poses significant challenges when their phase relations vary non-linearly with time. Given that DTW associates the time points of sequences based on their similarities to eliminate phase variation, several methods have been developed that use these associations in the computation of the mean of a collection of sequences. They aim at producing a mean sequence by minimizing the Within Group Sum of Squares (WGSS) in terms of its DTW distance from the averaged sequences. In this paper, we show that due to the nature of the DTW distance, which reduces temporal information to sequential, previous methods result in inexact averaging of the phase variation, and incorrect and unreliable estimation of the phase of the time points of the mean sequence. As such information is essential for many tasks where timing is a critical factor, we propose a novel method for computing the mean sequence that accurately averages phase variation. The method is founded on the definition of the Time Warp Profile that transforms the optimal warping path from a set of associations into a regularly sampled sequence with a predetermined length. Furthermore, TWP have applications beyond the averaging of time series. Based on TWP, we define a new and robust measure of the global phase difference between two sequences, the Phase distance. This enabled us to define the mean sequence as the "centre" of the phase variation that minimizes the WGSS with respect to the phase distance. Additionally, we were able to generalize the Time Alignment Measurement [26] for warping paths that may include any type of step pattern.

We performed qualitative and quantitative evaluations of our TWP time series averaging method and showed that it achieves accurate estimation of the ground truth in synthetic sequences as well as in real world data that contain a high degree of variation. In addition, we show that descriptive statistic quantities, such as the standard deviation around the mean, may also be accurately computed. In terms of standard datasets, we evaluated our method with datasets from the UCR archive. We showed that it performs better than the state-of-the-art in all datasets in terms of WGSS with respect to the relevant distance measures, the Phase and TAM distances, and in half of the datasets, with respect to the less relevant DTW distance. We conclude that our method offers a generic and efficient sequence averaging technique that accurately estimates and preserves the distinctive durational features of a collection of signals.

In terms of future work, we identify two main aspects of the averaging algorithm where further studies could improve the results. First, the resulting mean sequence may benefit from employing the recent DTW methods that improve the alignment between sequences. Such methods include the shapeDTW algorithm [9] that use feature similarity instead of similarity between individual time points. Second, alternative termination conditions and distance metrics between the sequences in a dataset may reduce the number of iterations required, and therefore the efficiency of the method, without degrading the quality of the outcome.

Furthermore, we may use TWP to create theoretically robust measures of warping paths' properties, similarly to how we defined the Phase distance and generalized TAM. Another interesting application of TWP for future exploration is the recognition of phase variation patterns derived from warping paths. For instance, TWP may substitute time sequences with distinctive temporal properties in motion classification tasks in sports or rehabilitation.

In conclusion, both the novel time series averaging method and the time warp profiles (TWP) presented in this paper overcome fundamental barriers of current methodologies and open major new possibilities in several fields of research and domains of application, such as information retrieval, time series summarization and signal noise suppression.

## 7    Acknowledgments

We would like to express our appreciation to Professor Anne Danielsen for her valuable feedback that greatly improved the manuscript.

**Funding**: This work was partially supported by the Research Council of Norway through its Centres of Excellence scheme, project number 262762.